\documentclass{bmvc2k}

\usepackage{txfonts}
\usepackage{mathrsfs}
\usepackage{amsmath}
\usepackage{graphicx}
\usepackage{footnote}
\def\infinity{\rotatebox{90}{8}}
\title{Fast Feature Fool: A data independent approach to universal adversarial perturbations}

\addauthor{Konda Reddy Mopuri$^\ast$}{sercmkreddy@grads.cds.iisc.ac.in}{1}
\addauthor{Utsav Garg$^{\ast\dagger}$}{utsav002@e.ntu.edu.sg}{2}
\addauthor{R. Venkatesh Babu}{venky@cds.iisc.ac.in}{1}
\addinstitution{
 Video Analytics Lab,\\
 Department of Computational and Data Sciences,\\
 Indian Institute of Science,\\
 Bangalore, India
}

\runninghead{Mopuri, Garg, Babu}{Fast Feature Fool}

\def\etal{\emph{et al}\bmvaOneDot}

\begin{document}

\maketitle

\begin{abstract}
State-of-the-art object recognition Convolutional Neural Networks (CNNs) are shown to be fooled by image agnostic perturbations, called universal adversarial perturbations. It is also observed that these perturbations generalize across multiple networks trained on the same target data. However, these algorithms require training data on which the CNNs were trained and compute adversarial perturbations via complex optimization. The fooling performance of these approaches is directly proportional to the amount of available training data. This makes them unsuitable for practical attacks since its unreasonable for an attacker to have access to the training data. In this paper, for the first time, we propose a novel data independent approach to generate image agnostic perturbations for a range of CNNs trained for object recognition. We further show that these perturbations are transferable across multiple network architectures trained either on same or different data. In the absence of data, our method generates universal perturbations efficiently via fooling the features learned at multiple layers thereby causing CNNs to misclassify. Experiments demonstrate impressive fooling rates and surprising transferability for the proposed universal perturbations generated without any training data.
\end{abstract}

\section{Introduction}
\label{sec:intro}
Machine learning systems are vulnerable~\cite{prsystemsunderattack-pari-2014,evasion-mlkd-2013,adversarialml-acmmm-2011} to adversarial samples - malicious input with structured perturbations that can fool the systems to infer wrong predictions. Recently, Deep Convolutional Neural Network (CNN) based object classifiers are also shown ~\cite{intriguing-arxiv-2013,explainingharnessing-arxiv-2014,deepfool-cvpr-2016,atscale-arxiv-2016, practicalbb-arxiv-2016} to be fooled by adversarial perturbations that are quasi-imperceptible to humans. There have been multiple approaches formulated to compute the adversarial samples, exploiting linearity of the models~\cite{explainingharnessing-arxiv-2014}, finite training data~\cite{deeparch-FTML-2009}, etc. More importantly, adversarial samples can be transferred (generalized) from one model to another, even if the second model has a different architecture and trained on different subset of training data~\cite{intriguing-arxiv-2013,explainingharnessing-arxiv-2014}. This property allows an attacker to launch an attack without the knowledge of the target model's internals, which makes them a dangerous threat for deploying the models in practice. Particularly, for critical applications that involve safety, robust models should be learned towards adversarial attacks. Therefore, the effect of adversarial perturbations warrants the need for in depth analysis of this subject.

Recent work by Moosavi-Dezfooli \etal \cite{universal-arxiv-2016} has shown that there exists a single perturbation image, called universal adversarial perturbation (UAP), that can fool a model with high probability when added to any data sample. These perturbations are image agnostic and show transferability (being able to fool) across multiple networks trained on the same data.

However, this method requires training data on which the target model is trained. They solve a complex data dependent optimization (equation~(\ref{eqn:existing})) to design a single perturbation that when added can flip the classifier's prediction for most of the images. Using a subset of this training data, they iteratively update the universal perturbation with the objective of changing the predicted label. It is observed that their optimization procedure requires certain minimum amount of training data in order to converge. Moreover, the fooling performance of these approaches are directly proportional to the amount of available training data (Figure~\ref{fig:data-comparison}). 
 This data dependence makes the approach not suitable for practical attacks as training data of the target system is generally unavailable.

In order to address these shortcomings, we propose a novel \textbf{\emph{data independent}} method to compute universal adversarial perturbations. The objective of our approach is to generate a single perturbation that can fool a target CNN on most of the images without any knowledge of the target data, such as, type of data distribution (eg: faces, objects, scenes, etc.), number of classes, sample images, etc. As our method has no access to data to learn a perturbation that can flip the classifier's label, we aim to fool the features learned by the CNN. In other words, we formulate this as an optimization problem to compute the perturbation which can fool the features learned at individual layers in a CNN and eventually making it to misclassify a perturbed sample. Our method is computationally efficient and can compute universal perturbation for any target CNN quickly (Section~\ref{subsec:comparison} and Table~\ref{tab:time-compare}), thereby named \textbf{\emph{Fast Feature Fool}} algorithm. The main contributions of our work are listed below

\begin{itemize}
    \item We introduce for the first time, a novel data independent approach to compute universal adversarial perturbations. To the best of our knowledge, there exists no previous work that can generate adversarial perturbations, universal or otherwise, without access to target training data. In fact, the proposed method doesn't require any knowledge about target data distribution, all that it requires is the target network.
    \item We show that misfiring the features learned at individual layers of a CNN to produce impartial (undiscriminating) activations can lead to eventual misclassification of the sample. We present an efficient and generic objective to construct image agnostic perturbations that can misfire the features and fool the CNN with high probability.
    \item We show that similar to data dependent universal perturbations, our data independent perturbations also exhibit remarkable transferability across multiple networks trained on same data. In addition to that, we show transferability across same architectures trained on different data. 
    As our method is explicitly made data independent, the transfer performance of our method is far better compared to that of data dependent methods (Section~\ref{subsec:transfer-data} and Table~\ref{tab:places-compare}). This property of data independent approach emphasizes the necessity for more research focus in this direction.
\end{itemize}

The paper is organized as following: section~\ref{sec:RelWorks} discusses existing approaches to compute adversarial perturbations, section~\ref{sec:Prop} presents the proposed data independent approach in detail, section~\ref{sec:Expts} demonstrates the effectiveness of the proposed method to fool CNNs and their transferability via a set of comprehensive experiments and section~\ref{sec:Conclu} hosts discussion and provides useful inferences 
and directions to continue further study about the adversarial perturbations and design of robust CNN classifiers.
\section{Related Works}
\label{sec:RelWorks}

Szegedy \etal~\cite{intriguing-arxiv-2013} observed that, despite their excellent recognition performances, neural networks get fooled by structured perturbations that are quasi-imperceptible to humans. Later, multiple other~\cite{robustnessanalysis-arxiv-2015, adversarial2noise-nips-2016,explainingharnessing-arxiv-2014, atscale-arxiv-2016,deepfool-cvpr-2016,easilyfooled-cvpr-2015,hardpositives-cvpr-2016} investigations studied this interesting property called, adversarial perturbations. These crafted malicious perturbations can be estimated per data point by simple gradient ascent~\cite{explainingharnessing-arxiv-2014} or complex optimizations~\cite{intriguing-arxiv-2013,deepfool-cvpr-2016}. Note that, the underlying property for all these methods is, they are intrinsically data dependent. The perturbations are computed for each data point specifically and independent of each other. 

Moosavi-Dezfooli \etal~\cite{universal-arxiv-2016} consider a generic problem to craft a single perturbation to fool a given CNN on most of the natural images, called universal perturbation. They collect samples from the data distribution and iteratively craft a single perturbation that can flip the labels over these samples. Fooling any new image now involves just an addition of the universal perturbation to it (no more optimization). In their work, they investigate the existence of a single adversarial direction in the space that is sufficient to fool most of the images from the data distribution. However, it is observed that the convergence of their optimization requires to sample enough data points from the distribution. Also, fooling rate increases proportionately with the sample size (Figure~\ref{fig:data-comparison}), making it an inefficient attack. 

Another line of research~\cite{limitations-eurosp-2016}, called oracle-based black box attacks, trains a local CNN with crafted inputs and output labels provided by the target CNN (called victim network). They use the local network to craft adversarial samples and show that they are effective on the victim (original) network also.

On the other hand, the proposed approach aims towards solving a more generic and difficult problem: we seek data independent universal perturbations, without sampling any images from the data distribution or train a local replica of the target model. We exploit the fundamental hierarchical nature of the features learned by the target CNN to fool it over most of the natural images. Also, we explore the existence of universal perturbations that can be transferred across architectures trained on different data distributions.
\section{Fast Feature Fool}
\label{sec:Prop}
In this section we present the proposed Fast Feature Fool algorithm to compute universal adversarial perturbations in a data independent fashion. 

First, we introduce the notation followed throughout the paper. Let \textbf{$\mathcal X$} denote the distribution of images in $\mathbb{R}^d$ and $f$ denotes the classification function learned by a CNN that maps an image $x \sim \mathcal X$ from the distribution to an estimated label $f(x)$.

The objective of this paper is to find a perturbation $\delta \in \mathbb{R}^d$ that fools the classifier $f$ on a large fraction of data points from \textbf{$\mathcal X$} without utilizing any samples. In other words, we seek a data independent universal (image agnostic) perturbation that can misclassify majority of the target data samples. That is, we construct a $\delta$ such that

\begin{equation}
    f(x+\delta) \neq f(x), \:\:\:\text{for majority of $x \in \mathcal X$}
\end{equation}
For the perturbation to be called adversarial, it has to be quasi-imperceptible for humans. That is, the pixel intensities of perturbation $\delta$ should be restricted. Existing works (eg:~\cite{adversarialmanipulation-arxiv-2015,universal-arxiv-2016,deepfool-cvpr-2016}) impose an $l_{\infinity}$ constraint $(\xi)$ on the perturbation $\delta$ to realize imperceptibility. Therefore, the goal here is to find a $\delta$ such that,
\begin{align}
\label{eqn:existing}
\begin{split}
  f(x+\delta) &\neq f(x), \:\:\:\text{for most $x \in \mathcal X$} \\
  ||\: \delta\: ||_{\infinity} &< \xi
\end{split}
\end{align}
As the primary focus of the proposed approach is to craft the universal perturbations ($\delta$) without any knowledge about target data \textbf{$\mathcal X$}, we attempt to exploit the dependencies across the layers in a given CNN. 
The data independence prohibits us to impose the first part of equation~\ref{eqn:existing} while learning $\delta$. Therefore, we propose to fool the CNN by over-saturating the features learned at multiple layers (replacing the ``flipping the label" objective). That is, by adding perturbation to input, we make the features at each layer to misfire thereby misleading the features (filters) at the following layer. This cumulative disturbance of features along the network hierarchy makes the network impartial (undiscriminating) to the original input, leading to an erroneous prediction in the final layer. 

The perturbation should essentially cause filters at a particular layer to spuriously fire and abstract out uninformative activations. Note that in the presence of data (during the attack), to mislead the activations from retaining the discriminative information, the perturbation has to be highly effective, given the added \emph{imperceptibility} constraint (second part of equation~\ref{eqn:existing}). Therefore, the difficulty of the attempted problem lies in crafting a perturbation $\delta$ whose dynamic range is restricted typically~\cite{adversarialmanipulation-arxiv-2015,universal-arxiv-2016,deepfool-cvpr-2016} (with $\xi =10$) to less than $8\%$ of the data range.

Hence, without showing any data $(x)$ to the target CNN, we seek for a perturbation $(\delta)$ that can produce maximal spurious activations at each layer. In order to obtain such a perturbation, we start with a random $\delta$ and optimize for the following loss
\begin{align}
\label{eqn:fff-prod}
\begin{split}
  Loss = - \log \left( \prod\limits_{i=1}^K  \bar l_i(\delta) \right) \:\:\:\:\:\: \text{such that} \:\:\:\:\:\:
  ||\: \delta\: ||_{\infinity} < \xi
\end{split}
\end{align}

where, $\bar l_i(\delta)$ is the mean activation in the output tensor at layer $i$ when $\delta$ is input to the CNN. 
Note that the activations are considered after the non-linearity (typically ReLU), therefore $\bar l_i$ is non-negative. $K$ is the total number of layers in the CNN at which we maximize activations for the perturbation $\delta$. We typically consider all the convolution layers before the fully connected layers (see section~\ref{sec:Expts} for advanced architectures). This is because, the convolution layers are generally considered to learn suitable features to extract information over which a series of fully connected layers act as a classifier. Also, we empirically found that it is sufficient to optimize at convolution layers. Therefore, we restrict the optimization to feature extraction layers, $\xi$ is the limit on the pixel intensity of the perturbation $\delta$. 

The proposed objective computes product of mean activations at multiple layers in order to simultaneously maximize the perturbation at all those layers. We observed that product results in a stronger $\delta$ than other forms of combining the individual layer activations (eg: sum). This is understandable, as product is a stronger constraint that forces activations at all layers to increase for the loss to reduce. To avoid working with extreme values ($\approx 0$), we apply $log$ on the product. Note that the objective is open-ended as there is no optimum value to reach. We would ideally want $\delta$ to cause as much strong perturbation at all the layers as possible within the imperceptibility constraint. 

We begin with a trained network and a random perturbation image $\delta$. We then perform the above optimization to update $\delta$ to achieve higher activations at all the convolution layers in the given network. Note that the optimization updates the perturbation image $\delta$ but not the network parameters and no image data is involved in the optimization. We update  $\delta$ with the gradients computed for loss in equation~(\ref{eqn:fff-prod}). After every update step, we clip the perturbation $\delta$ to satisfy the imperceptibility constraint. We treat the algorithm has converged when either the loss gets saturated or fooling performance over a small held out set is maximized.

\section{Experiments}
\label{sec:Expts}
In this section, we evaluate the proposed Fast Feature Fool method to fool multiple CNN architectures trained on ILSVRC~\cite{imagenet-ijcv-2015} dataset. Particularly, we considered CaffeNet (similar to Alexnet~\cite{deepcnn-nips-2012}), VGG-F~\cite{vggf-bmvc-2014}, VGG-16~\cite{vgg-arxiv-2014}, VGG-19~\cite{vgg-arxiv-2014} and GoogLeNet~\cite{googlenet-arxiv-21014} architectures. As the primary experiment, we compute the image agnostic universal perturbations for each of these CNNs via optimizing the loss given in equation~(\ref{eqn:fff-prod}). 
For simple networks (eg: CaffeNet) we optimize the activations at all the convolution layers after the non-linearity (ReLU). However, for networks with complex architectures (inception block) such as GoogLeNet, we optimize activations at selected layers. Specifically, for GoogLeNet, we compute the perturbations by maximizing the activations at all the concat layers of inception blocks and conv layers that are not part of any inception block. 
This is because maximizing at concat layer for the case of inception blocks inherently takes care of optimization for the convolution layers since they are part of concat layers. 

\begin{figure}[h]
\centering
\noindent\begin{minipage}{\textwidth}
  \centering
  \begin{minipage}{.19\textwidth}
  	\centering
    \includegraphics[width=\linewidth]{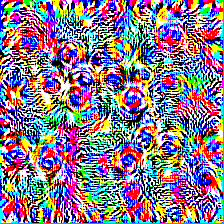}\
    VGG-F
  \end{minipage}
   \begin{minipage}{.19\textwidth}
   	\centering
    \includegraphics[width=\linewidth]{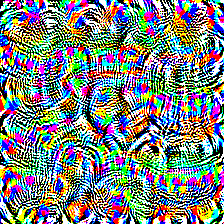}\
    CaffeNet
  \end{minipage}
  \begin{minipage}{.19\textwidth}
  	\centering
    \includegraphics[width=\linewidth]{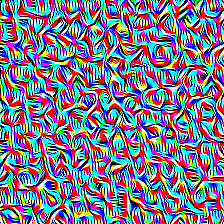}\
    GoogLeNet
  \end{minipage}
  \begin{minipage}{.19\textwidth}
  \centering
    \includegraphics[width=\linewidth]{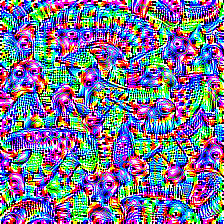}\
    VGG-16
  \end{minipage}
   \begin{minipage}{.19\textwidth}
   \centering
    \includegraphics[width=\linewidth]{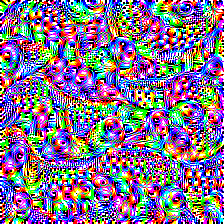}\
    VGG-19
  \end{minipage}
    \vspace{0.002\textwidth}
\end{minipage}
\caption{Data independent universal adversarial perturbations crafted by the proposed method for multiple architectures trained on ILSVRC~\cite{imagenet-ijcv-2015} dataset. Perturbations were crafted with $\xi=10$. Corresponding target network architecture is mentioned below each image. Images are best viewed in color.}
\label{fig:ourPerturbations}
\end{figure}

\begin{figure}[h]
\centering
\noindent\begin{minipage}{\textwidth}
  \centering
  \begin{minipage}{.15\textwidth}
  	\centering
    \includegraphics[width=\linewidth]{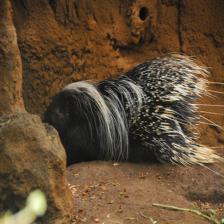}\
    marmoset
  \end{minipage}
   \begin{minipage}{.15\textwidth}
   	\centering
    \includegraphics[width=\linewidth]{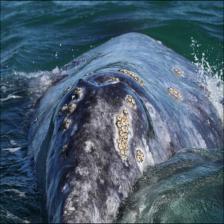}\
    grey whale
  \end{minipage}
  \begin{minipage}{.15\textwidth}
  	\centering
    \includegraphics[width=\linewidth]{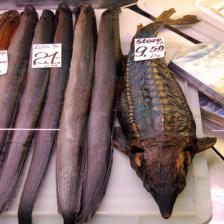}\
    sturgeon
  \end{minipage}
  \begin{minipage}{.15\textwidth}
  	\centering
    \includegraphics[width=\linewidth]{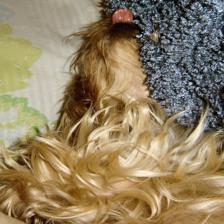}\
    wig
  \end{minipage}
   \begin{minipage}{.15\textwidth}
   	\centering
    \includegraphics[width=\linewidth]{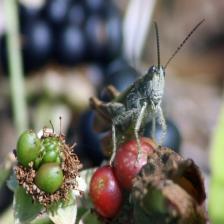}\
    acorn
  \end{minipage}
  \begin{minipage}{.15\textwidth}
  	\centering
    \includegraphics[width=\linewidth]{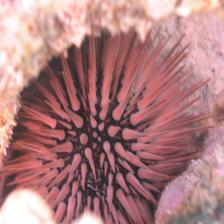}\
     sea urchin
  \end{minipage}
  \vspace{0.002\textwidth}
\end{minipage}
\noindent\begin{minipage}{\textwidth}
  \centering
  \begin{minipage}{.15\textwidth}
  \centering
    \includegraphics[width=\linewidth]{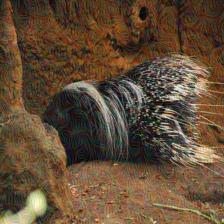}\
    tabby
  \end{minipage}
   \begin{minipage}{.15\textwidth}
   \centering
    \includegraphics[width=\linewidth]{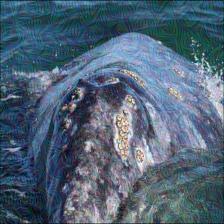}\
    turtle
  \end{minipage}
  \begin{minipage}{.15\textwidth}
  	\centering
    \includegraphics[width=\linewidth]{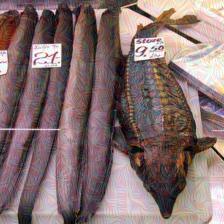}\
    butcher shop
  \end{minipage}
  \begin{minipage}{.15\textwidth}
  \centering
    \includegraphics[width=\linewidth]{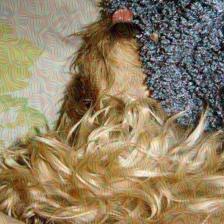}\
     terrier
  \end{minipage}
   \begin{minipage}{.15\textwidth}
   \centering
    \includegraphics[width=\linewidth]{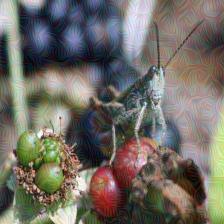}\
    tarantula
  \end{minipage}
  \begin{minipage}{.15\textwidth}
  	\centering
    \includegraphics[width=\linewidth]{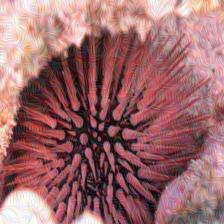}\
    wool
  \end{minipage}
    \vspace{0.002\textwidth}
\end{minipage}
\vspace{0.1cm}
\caption{Sample original and adversarial image pairs from ILSVRC validation set generated for GoogLeNet. First row shows original images and corresponding predicted labels, second row shows the corresponding perturbed images along with their predictions. Note that all of the shown perturbed images were misclassified.}
\label{fig:ourPerturbedImages}
\end{figure}

Similar to existing approaches~\cite{deepfool-cvpr-2016,universal-arxiv-2016,adversarialmanipulation-arxiv-2015}, we restricted the pixel intensities of the perturbation to lie within $\big[-10, +10\big]$ range by choosing $\xi$ in equation~(\ref{eqn:fff-prod}) to be $10$. Figure~\ref{fig:ourPerturbations} shows the universal perturbations $\delta$ obtained for the networks using the proposed method. Note that the perturbations are visually different for each network architecture. Figure~\ref{fig:ourPerturbedImages} shows sample perturbed images $(x+\delta)$ for GoogLenet from ILSVRC validation set along with their corresponding original images. Note that the adversarial images are perceptually indistinguishable from original ones and yet get misclassified by the CNN.
\subsection{Transferability across network architectures}
\label{subsec:transfer-nets}
Interesting property of the proposed data independent universal perturbations is that they transfer not only to other images but also to different network architectures trained on the same dataset. That is, we compute universal perturbation on one architecture (eg: VGG-F) and observe its ability to fool on other networks (eg: GoogLeNet). This property is observed in the case of existing data dependent perturbations~\cite{universal-arxiv-2016,intriguing-arxiv-2013,explainingharnessing-arxiv-2014} as well. However, transferability of proposed perturbations is a serious issue and needs more attention, as the perturbation is crafted without any data.
\begin{table}[t]
\centering
\caption{Fooling rates for the proposed perturbations crafted for multiple networks trained on ILSVRC dataset computed over $50000$ validation images. Each row shows fooling rates for perturbation crafted for a particular network. Each column shows the transfer fooling rates obtained for a given network. Diagonal values in bold are the fooling rates obtained via dedicated optimization for each architecture.}
\label{tab:transfer-nets}
\begin{tabular}{l|c|c|c|c|c|}
\cline{2-6}
\textbf{}                                & \textbf{VGG-F}   & \textbf{CaffeNet} & \textbf{GoogLeNet} & \textbf{VGG-16}  & \textbf{VGG-19}  \\ \hline
\multicolumn{1}{|l|}{\textbf{VGG-F}}     & \textbf{81.59\%} & 48.20\%           & 38.56\%            & 39.31\%          & 39.19\%          \\ \hline
\multicolumn{1}{|l|}{\textbf{CaffeNet}}  & 56.18\%          & \textbf{80.92\%}  & 39.38\%            & 37.22\%          & 37.62\%          \\ \hline
\multicolumn{1}{|l|}{\textbf{GoogLeNet}} & 49.73\%          & 46.84\%           & \textbf{56.44\%}   & 40.91\%          & 40.17\%          \\ \hline
\multicolumn{1}{|l|}{\textbf{VGG-16}}    & 46.49\%          & 43.31\%           & 34.33\%            & \textbf{47.10\%} & 41.98\%          \\ \hline
\multicolumn{1}{|l|}{\textbf{VGG-19}}    & 39.91\%          & 37.95\%           & 30.71\%            & 38.19\%          & \textbf{43.62\%} \\ \hline
\end{tabular}
\label{tab:ourFoolingRates}
\end{table}
Table~\ref{tab:transfer-nets} presents the fooling rates (\% of images for which the predicted label is flipped by adding $\delta$) of the proposed approach including the transfer rates on multiple networks. 
Note that all these architectures are trained on ILSVRC dataset and the fooling rates are computed for the $50000$ validation images from the dataset. Each row in Table~\ref{tab:transfer-nets} shows fooling rates for perturbation crafted for a particular network. Each column shows the transfer fooling rates obtained for a given network. Diagonal values in bold are the fooling rates obtained for dedicated optimization for each of the architectures. Observe that, the perturbations crafted for some of the architectures generalize very well across multiple networks. For example, perturbation obtained for GoogLeNet ($3^{rd}$ row in Table~\ref{tab:transfer-nets}) has a minimum fooling rate of $40.17\%$ across all the tested networks. 
Note that the average transfer rate of the crafted perturbations for all the five networks is $41.31\%$ which is very significant given the method has no knowledge about target data distribution. These results show that our data independent universal perturbations generalize well across different architectures and are of practical importance.
\subsection{Transferability across data}
\label{subsec:transfer-data}
Until now, existing methods have shown the transferability of adversarial perturbations to images belonging to single distribution (eg: ILSVRC dataset) and multiple networks trained on same target distribution (some times on disjoint subsets~\cite{intriguing-arxiv-2013}). We now extend the notion of transferability by considering multiple data distributions. That is, we compute universal perturbation for a given CNN trained over one dataset (eg: ILSVRC~\cite{imagenet-ijcv-2015}) and observe the fooling rate for the same architecture trained on another dataset (eg: Places~\cite{places2-arxiv-2016}). For this evaluation, we have considered three network architectures trained on ILSVRC and Places-205 datasets. Since the proposed objective is data independent and the crafted perturbations aim at fooling the learned features, we study the extent to which the perturbations fool similar filters learned on different data. Table~\ref{tab:places-compare} presents the change in fooling rates for the proposed method when evaluated across datasets for multiple networks. 

\begin{table}[h]
\centering
\caption{Comparing transferability across data. Change in fooling rates for the perturbations crafted for architectures trained on ILSVRC~\cite{imagenet-ijcv-2015} and evaluated on same architectures trained on Places-205~\cite{places2-arxiv-2016}. The results clearly show that the absolute change in fooling rate for UAP~\cite{universal-arxiv-2016} is significantly higher than our approach because of the strong data dependence. Note that the perturbation trained on CaffeNet was tested on AlexNet trained on Places (the networks differ slightly) and explains the larger drop in case of CaffeNet for the proposed approach.}
\label{tab:places-compare}
\begin{tabular}{l|c|c|c|}
\cline{2-4}
\textbf{}                           & \textbf{CaffeNet*} & \textbf{VGG-16}  & \textbf{GoogLeNet} \\ \hline
\multicolumn{1}{|l|}{\textbf{UAP}}  & 30.05\%          & 18.89\%         & 24.50\%           \\ \hline
\multicolumn{1}{|l|}{\textbf{Ours}} & \textbf{18.59\%} & \textbf{3.96\%} & \textbf{6.10\%}    \\ \hline
\end{tabular}
\end{table}

In order to bring out the effectiveness of the proposed data independent perturbations, we compared with the performance of data dependent perturbations (UAP)~\cite{universal-arxiv-2016}. We have evaluated the fooling rates of~\cite{universal-arxiv-2016}, crafted on ILSVRC and transferred to Places-$205$. The validation set of Places-205 dataset contains $20500$ images from $205$ scene categories over which the fooling rates are computed. Table~\ref{tab:places-compare} shows the absolute change in fooling rate ( $|$ rate on ILSVRC - rate on Places-$205$ $|$ ) of the perturbations when evaluated on Places-205 trained architectures. It is clearly observed that the proposed approach on average suffers less change in the fooling rate compared to data dependent approach. 
Note that the data dependent universal perturbations quickly lose their ability to fool the features trained on different data even for the same architecture. This is understandable as they are crafted in association with target data distribution (\textbf{$\mathcal X$}). Unlike the data dependent approaches, the proposed perturbations are not tied to any target data and the objective makes them more generic to fool data from different distributions.
\subsection{Initialization with smaller network's perturbation}
\label{subsec:init}
\vspace{-.1cm}
In all the above set of experiments we begin our optimization with perturbation initialized with uniform random distribution $\big[-10, +10\big]$. In this section, we investigate the effect of a pretrained initialization for $\delta$ on the proposed objective and optimization. We consider perturbations computed for shallow network (eg: VGG-F) as initialization and optimize for deeper nets (GoogLeNet, VGG-16, VGG-19). Note that all the networks are trained on ILSVRC dataset. Table~\ref{tab:init} shows the obtained fooling rates for other networks when initialized with VGG-F's perturbation, and shows the improvement over random initialization(shown in parentheses). This improvement is understandable as the perturbation from VGG-F already has some structure and shows transferability on the deeper networks, therefore optimization using this offers slight imporvement when compared to random initialization.
\begin{table}[h]
\centering
\caption{Fooling rates for deeper networks initialized with smaller network's perturbation. All the networks are trained on ILSVRC dataset. Proposed universal perturbations are computed with VGG-F's perturbation as initialization. Improvent over random initialization shown in parenthesis.}
\label{tab:init}
\begin{tabular}{|l|c|c|c|c|}
\hline
\textbf{Network}        &  \textbf{GoogLeNet} & \textbf{VGG-16} & \textbf{VGG-19} \\ \hline
\textbf{Fooling Rate}   & 58.93\% (2.49)           & 49.47\% (2.37)         & 45.87\% (2.25)            \\ \hline
\end{tabular}
\end{table}
\subsection{Comparison with data dependent universal perturbations}
\label{subsec:comparison}
In this section, we investigate  how the proposed perturbations compare to data dependent counterparts. We consider two cases, when the data dependent approach (UAP\cite{universal-arxiv-2016}): (i) has access to the target dataset, and (ii) uses images form a different dataset instead of the target dataset.
\subsubsection{With access to target dataset}
\begin{figure}[h]
    \centering
    \includegraphics[width=.75\linewidth]{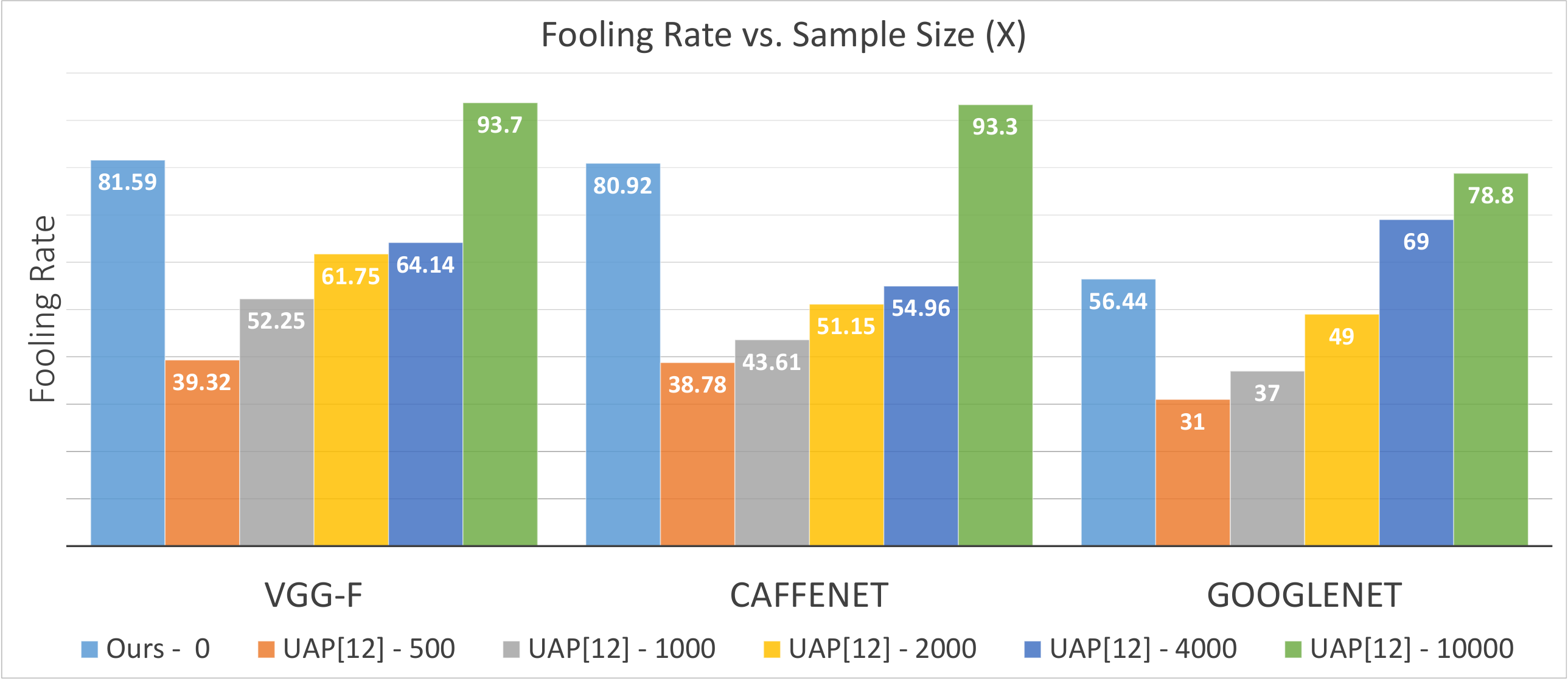}
    \caption{Comparision of fooling rates for the proposed approach with UAP~\cite{universal-arxiv-2016} for multiple networks trained on ILSVRC~\cite{imagenet-ijcv-2015}. Legend shows the size of (X) sampled from training data. Note that, due to their strong data dependence the performance of UAP~\cite{universal-arxiv-2016} increases monotonically with size of X for all networks. 
    This strong data dependence explains the larger drop in performance of UAP when tested on the same architecture trained on different datasets, as shown in Table~\ref{tab:places-compare}.}
    \label{fig:data-comparison}
\end{figure}

Note that the data dependent methods~\cite{universal-arxiv-2016} when utilize samples $(X)$ from target data distribution to craft the perturbations, they are expected to demonstrate higher fooling rates on the target datasets. However, we compare the fooling rates for different sample sizes $(X)$ versus our \emph{\textbf{no data}} fooling rate performance. Figure~\ref{fig:data-comparison} presents the comparison for three networks trained on ILSVRC dataset evaluated on $50000$ validation images. As evaluated in~\cite{universal-arxiv-2016}, we have computed fooling performance for $500, 1000, 2000$, $4000$ and $10000$ samples from the training data for them. Note that the fooling performance of~\cite{universal-arxiv-2016} improves monotonically with the sample $(X)$ size for all the networks. It shows the strong association of their perturbations to target data distribution and lose their ability to fool when evaluated on different data though the target architectures are similar. On the other hand, the proposed perturbations, as they do not utilize any data, they fool networks trained on other data equally well (Table~\ref{tab:places-compare}). 
\subsubsection{Without the access to target dataset}
While it is not a reasonable assumption to have access to the training data, we can argue that the data dependent methods (UAP) can use images from an arbitrary dataset and train the perturbations. In this section, we investigate by learning UAPs for a target architecture using an arbitrary dataset. We have considered ILSVRC and Places-205 datasets. 

 Table~\ref{tab:p2i} shows the fooling rates on ILSVRC trained networks, where we used Places-205 data to generate the perturbation for UAP~\cite{universal-arxiv-2016} and Table~\ref{tab:i2p} shows the reverse scenario. Note that in both cases our approach just needs the target network and no data. The numbers are computed on the validation set of the corresponding datasets. These experiments clearly show that the data dependent perturbations~\cite{universal-arxiv-2016}  are strongly tied to the target dataset and experience a significant drop in performance if the same is unavailable. It is also seen that this drop is more severe with larger networks (GoogLeNet). On the contrary, our approach without using any data results in significantly better performance for CNNs trained on both datasets. 
 \begin{table*}[h]
    \centering
    \begin{minipage}{.48\textwidth}
        \centering
        \caption{Fooling rates obtained when UAPs~\cite{universal-arxiv-2016} are trained and tested for ILSVRC architectures using the data from Places-205. Note that our approach doesn't require any data.}
        \label{tab:p2i}
        \begin{tabular}{|l|c|c|}
        \hline
                  & \textbf{Ours}  & \textbf{UAP~\cite{universal-arxiv-2016}}   \\ \hline
        \textbf{CaffeNet}  & \textbf{80.92\%} & 73.09\%                                           \\ \hline
        \textbf{GoogleNet} & \textbf{56.44\%} & 28.17\%                                           \\ \hline
        \end{tabular}       
    \end{minipage}%
    \hfill
    \begin{minipage}{0.48\textwidth}
        \centering
        \caption{Fooling rates obtained when UAPs~\cite{universal-arxiv-2016} are trained and tested for Places-205 architectures using the data from ILSVRC. Note that our approach doesn't require any data.}
        \label{tab:i2p}
        \begin{tabular}{|l|c|c|}
        \hline
                  & \textbf{Ours}  & \textbf{UAP~\cite{universal-arxiv-2016}}   \\ \hline
        \textbf{CaffeNet}  & \textbf{87.61\%} & 77.21\%                                          \\ \hline
        \textbf{GoogleNet} & \textbf{78.08\%} & 52.53\%                                           \\ \hline
        \end{tabular}
    \end{minipage}
\end{table*}
\subsubsection{Convergence time}
The time of convergence for the proposed optimization is compared with that of data dependent universal perturbations approach~\cite{universal-arxiv-2016} in Table~\ref{tab:time-compare}. We have utilized the implementation provided by the authors of~\cite{universal-arxiv-2016} that samples $10000$ training images. Convergence time is reported in seconds for both the approaches on three different network architectures trained on ILSVRC dataset. Note that the proposed approach takes only a small fraction of time taken by~\cite{universal-arxiv-2016}. We have run the timing experiments on an NVIDIA GeForce TITAN-X GPU with no other jobs on the system.
\begin{table}[]
\centering
\caption{Comparison of the time of convergence for UAP~\cite{universal-arxiv-2016} and the proposed approch. It is observed that the proposed approach takes only a fraction of time compared to UAP across different network architechtures.}
\label{tab:time-compare}
\begin{tabular}{l|c|c|c|}
\cline{2-4}
\textbf{}                           & \textbf{VGG-F}  & \textbf{CaffeNet} & \textbf{GoogLeNet} \\ \hline
\multicolumn{1}{|l|}{\textbf{UAP~\cite{universal-arxiv-2016}}}  & 3507.33s        & 3947.36s          & 13780.72s          \\ \hline
\multicolumn{1}{|l|}{\textbf{Ours}} & \textbf{69.49s} & \textbf{54.37s}   & \textbf{127.66s}   \\ \hline
\end{tabular}
\end{table}
\subsection{Implementation details}
\label{subsec:Imple}
In this section, for the ease of reproducibility we explain the implementation details of the proposed approach. We conducted all experiments using the TensorFlow~\cite{tensorflow2015-whitepaper-short} framework. As the objective is to craft universal perturbations, for each network we extracted the activations at all convolution or concat (for inception) layers and formulated the loss as the \emph{log} of product of activations at different layers~(eq \ref{eqn:fff-prod}), we minimized the negative of this and so the loss is unbounded in the negative direction. We used the Adam~\cite{kingma2014adam} optimizer with a learning rate of $0.1$ with other parameters set at their default values. We monitor the loss to see when it saturates and check validation over a held out set of $1000$ images to save the perturbation. Since, the optimization updates just the input ($\delta$) which is restricted to [-10 10] range, it gets saturated very quickly. Therefore, to avoid later updates from being ignored we periodically rescale the perturbation to [-5 5] range and then continue the optimization. Empirically we found rescaling at every $300$ iterations to work better and we use that for all our experiments. Project codes are available at \url{https://github.com/utsavgarg/fast-feature-fool}.

\section{Conclusion}
\label{sec:Conclu}
We have presented a simple and effective procedure to compute data independent universal adversarial perturbations. The proposed perturbations are quasi-imperceptible to humans but they fool state-of-the-art CNNs with significantly high fooling rates. In fact, the proposed perturbations are triply universal: (i) the same perturbation can fool multiple images form the target dataset over a given CNN, (ii) they demonstrate transferability across multiple networks trained on same dataset and (iii) they surprisingly retain (compared to data dependent perturbations) the ability to fool CNNs trained on different target dataset. 

Experiments (sections~\ref{subsec:transfer-nets} and~\ref{subsec:transfer-data}) demonstrate that data independent universal adversarial perturbations can pose a more serious threat when compared to their data dependent counterparts. They can enable the attackers not to be concerned about either the dataset on which the target models are trained or the internals of the model themselves. At this point in time, a more rigorous study (in case of extreme depth, presence of advanced regularizers, etc.) about the data independent aspect of the adversarial perturbations is of utmost importance. It should also be complemented simultaneously with the efforts to develop methods to learn more robust models. However, we believe our work opens new avenues into the intriguing aspects of adversarial machine learning with a data independent perspective.

\bibliography{mybibliography}
\end{document}